\documentclass[conference,letterpaper]{IEEEtran}

\usepackage[utf8]{inputenc} 
\usepackage[T1]{fontenc}
\usepackage{url}
\usepackage{ifthen}
\usepackage{cite}
\usepackage[cmex10]{amsmath} 

\usepackage[shortlabels]{enumitem}
\usepackage{graphicx}
\usepackage{epsfig}
\usepackage{times}
\usepackage{amssymb}
\usepackage{amsthm}
\usepackage{color}
\usepackage{float}
\usepackage{multirow}
\usepackage{color}
\renewcommand{\baselinestretch}{1}
\usepackage{enumitem}
\usepackage[normalem]{ulem}
\usepackage{url}
\usepackage{amsthm}
\usepackage{wrapfig}
\usepackage{hyperref}
\usepackage{mathtools}
\usepackage{mathnotations}

\usepackage{titlesec}

\newcommand{\prob}[1]{\mathbb{P}\left[{#1}\right]}
\newcommand{\given}{\; \big\vert \;}

\DeclareMathOperator{\F}{F}

\DeclareMathOperator{\D}{D}
\DeclareMathOperator{\diff}{d}
\DeclareMathOperator{\LL}{L}

\newcommand{\beq}{\begin{equation}}
\newcommand{\eeq}{\end{equation}}
\newcommand{\beqn}{\begin{equation*}}
\newcommand{\eeqn}{\end{equation*}}
\newcommand{\beqa}{\begin{eqnarray}}
\newcommand{\eeqa}{\end{eqnarray}}
\newcommand{\beqan}{\begin{eqnarray*}}
\newcommand{\eeqan}{\end{eqnarray*}}

\newcommand{\norm}[1]{\left\lVert #1 \right\rVert}
\DeclareMathOperator*{\argmin}{arg\,min} %

\def\tp{\widetilde{P}}

\newcommand{\ignore}[1]{}

\newcommand{\ex}[1]{\mathbb{E}\left[ #1 \right] }
\setlist[description]{
  leftmargin=\parindent, itemindent=-.5em,
  itemsep=-0.4em, topsep=.5em
}

\makeatletter
\def\thm@space@setup{\thm@preskip=2pt
\thm@postskip=0pt}
\makeatother

\newtheorem{thm}{Theorem}
\newtheorem{thm*}{Theorem}[section]

\newtheorem{cor*}{Corollary}[section]
\newtheorem{definition}{Definition}

\newtheorem*{corollary*}{Corollary}
\newtheorem{lemma}{Lemma}

\newtheorem*{lemma*}{Lemma}

\newtheorem*{prop*}{Proposition}

\newtheorem{assumption}{Assumption}

\usepackage[ruled,linesnumbered,vlined]{algorithm2e}
\usepackage{algpseudocode} 
\SetKwInOut{Parameter}{parameter}

\newcommand{\PRL}{\ensuremath{\mathrm{Private}\text{-}\mathrm{OFU}\text{-}\mathrm{RL}}}

\newcommand{\RL}{\ensuremath{\mathrm{OFU}\text{-}\mathrm{RL}}}
\newcommand{\psum}{\ensuremath{\mathrm{P}\text{-}\mathrm{sum}}}
\newcommand{\psums}{\ensuremath{\mathrm{P}\text{-}\mathrm{sums}}}

\titlespacing*{\section}{0pt}{0.75\baselineskip}{0.25\baselineskip}
\titlespacing*{\subsection}{0pt}{0.5\baselineskip}{0.25\baselineskip}

\begin{document}

\setlength{\belowdisplayskip}{3pt} \setlength{\belowdisplayshortskip}{3pt}
\setlength{\abovedisplayskip}{3pt} \setlength{\abovedisplayshortskip}{3pt}

\title{Adaptive Control of Differentially Private Linear Quadratic Systems} 


\author{%
  \IEEEauthorblockN{Sayak Ray Chowdhury\IEEEauthorrefmark{1}}
  \IEEEauthorblockA{Indian Institute of Science\\
                    Bangalore, India\\
                    Email: sayak@iisc.ac.in}
  \and
  \IEEEauthorblockN{Xingyu Zhou\IEEEauthorrefmark{1}}
  \IEEEauthorblockA{Wayne State University\\
                    Detroit, USA \\ 
                    Email: xingyu.zhou@wayne.edu}
   \and
  \IEEEauthorblockN{Ness Shroff}
  \IEEEauthorblockA{The Ohio State University\\
                    Columbus, USA \\ 
                    Email: shroff.11@osu.edu}
}


\maketitle
\begingroup\renewcommand\thefootnote{\IEEEauthorrefmark{1}}
\footnotetext{Equal contribution. This work was funded in part through NSF grants: CNS-1901057 and CNS- 2007231, and an Office of Naval Research under Grant N00014-17-1-241}
\endgroup
\begin{abstract}
In this paper we study the problem of regret minimization in reinforcement learning (RL) under differential privacy constraints. This work is motivated by the wide range of RL applications for providing personalized service, where privacy concerns are becoming paramount.
 In contrast to previous works, we take the first step towards \emph{non-tabular} RL settings, while providing a rigorous privacy guarantee. In particular, we consider the adaptive control of differentially private linear quadratic (LQ) systems. We develop the first private RL algorithm, \PRL\, which is able to attain a sub-linear regret while guaranteeing privacy protection. More importantly, the additional cost due to privacy is only on the order of $\frac{\ln(1/\delta)^{1/4}}{\epsilon^{1/2}}$ given privacy parameters $\epsilon, \delta > 0$. Through this process, we also provide a general procedure for adaptive control of LQ systems under \emph{changing regularizers}, which not only generalizes previous non-private controls, but also serves as the basis for general private controls.
\end{abstract}

\section{Introduction}
Reinforcement learning (RL) is a control-theoretic problem, which adaptively learns to make sequential decisions in an unknown environment through trial and error. 
RL has shown to have significant success for delivering a wide variety of personalized services, including online news and advertisement recommendation~\cite{li2010contextual}, 
medical treatment design~\cite{zhao2009reinforcement}, natural language processing~\cite{sharma2017literature}, and social robot~\cite{gordon2016affective}. 
In these applications, an RL agent improves its personalization algorithm by interacting with users to maximize the reward. In particular, in each round, the RL agent offers an action based on the user's state, and then receives the feedback from the user (i.e., state information, state transition, reward, etc.). This feedback is used by the agent to learn the unknown environment and improve its action selection strategy.

However, in most practical scenarios,  the feedback from the users often encodes their sensitive information. For example, in a personalized healthcare setting, the states of a patient include personal information such as age, gender, height, weight, state of the treatment etc. Similarly, the states of a virtual keyboard user (e.g., Google GBoard users) are the words and sentences she typed in, which inevitably contain private information about the user. Another intriguing example is the social robot for second language education of children. The states include facial expressions, and the rewards contain whether they have passed the quiz. Users may not want any of this information to be inferred by others. 
This directly results in an increasing concern about privacy protection in personalized services. To be more specific, although a user might be willing to share her own information to the agent to obtain a better tailored service, she would not like to allow third parties to infer her private information from the output of the learning algorithm. For example, in the healthcare application, we would like to ensure that an adversary with arbitrary side knowledge cannot infer  a particular patient's state from the treatments prescribed to her.

\emph{Differential privacy} (DP) \cite{dwork2008differential} has become a standard mechanism for designing interactive learning algorithms under a rigorous privacy guarantee for individual data. Most of the previous works on differentially private learning under partial feedback focus on the simpler bandit setting (i.e., no state transition) \cite{tossou2016algorithms,tossou2017achieving,basu2019differential,mishra2015nearly,zhou2020local}. For the general RL problem, there are only a few works that consider differential privacy \cite{balle2016differentially,vietri2020private,garcelon2020local}. More importantly, only the \emph{tabula-rasa} discrete-state discrete-action environments are considered in these works. However, in real-world applications mentioned above, the number of states and actions are often very large and can even be infinite. Over the years, for various non-tabular environments,  efficient and provably optimal algorithms for \emph{reward maximization} or, equivalently, \emph{regret minimization} have been developed (see, e.g., \cite{abbasi2011regret,osband2014model,chowdhury2019online,wang2020episodic,jin2020provably}).
This directly motivates the following question: \emph{Is it possible to obtain the optimal reward while providing individual privacy guarantees in the non-tabular RL scenario?}

In this paper, we take the first step to answer the aforementioned question by considering a particular non-tabular RL problem -- adaptive control of linear quadratic (LQ) systems, in which the state transition is a linear function and the immediate reward (cost) is a quadratic function of the current state and action. In particular, our main contributions can be summarized as follows.


\begin{itemize}
    \item First, we provide a general framework for adaptive control of LQ systems under \emph{changing regularizers} using the optimism in the face of uncertainty (OFU) principle, which covers both the extreme cases -- non-private and fully private LQ control.
    \item We then develop the first private RL algorithm, namely \PRL , for regret minimization in LQ systems by adapting the \emph{binary counting mechanism} to ensure differencial privacy.
    \item In particular, we show that \PRL\ satisfies \emph{joint differential privacy} (JDP), which, informally, implies that sensitive information about a given user is protected even if an adversary has access to the actions prescribed to all other users.
    \item Finally, we prove that \PRL\ achieves a sub-linear regret guarantee, where the regret due to privacy only grows as $\frac{\ln(1/\delta)^{1/4}}{\epsilon^{1/2}}$ with privacy levels $\epsilon,\delta \!>\! 0$ implying that a high amount of privacy (low $\epsilon,\delta$) comes at a high cost and vice-versa. 
\end{itemize}


\section{Preliminaries}

\subsection{Stochastic Linear Quadratic Control}
We consider the discrete-time episodic linear quadratic (LQ) control problem with $H$ time steps at every episode.
Let $x_{h} \!\in\! \Real^n$ be the state of the system, $u_{h} \!\in\! \Real^d$ be the control and $c_{h} \!\in\! \Real$ be the cost at time $h$. An LQ problem is characterized by linear dynamics and a quadratic cost function
\begin{equation}
x_{h+1} \!=\! A x_{h} + B u_{h} + w_{h},\;\;
    c_{h} \!=\! x_{h}^\top Q x_{h} + u_{h}^\top R u_{h}~,
\label{eq:model}
\end{equation}
where $A$, $B$ are \emph{unknown} matrices, and $Q$, $R$ are known positive definite (p.d.) matrices. The starting state $x_{1}$ is fixed (can possibly be chosen by an adversary) and the system noise $w_{h} \!\in\! \Real^n$ is zero-mean.  We summarize the unknown parameters in $\Theta \!=\! [A,B]^\top \!\in\! \Real^{(n+d)\times n}$ .


The goal of the agent is to design a closed-loop control
policy $\pi: [H] \times \Real^n \to \Real^d$ mapping states to controls that minimizes the expected cost
\begin{align}
\label{eq:pi_cost}
    J_{h}^{\pi}(\Theta,x) := \mathbb{E}_{\pi}\Big[\sum\nolimits_{h'=h}^H c_{h'} \given x_{h} = x\Big],
\end{align}
for all $h\in [H]$ and $x \in \Real^n$. Here the expectation is over the random trajectory induced by the policy $\pi$ starting from state $x$ at time $h$.
From the standard theory for LQ control (e.g.,~\cite{BertsekasDP04}), the optimal policy $\pi^*$ has the form
\begin{align*}
    \pi_{h}^*(x) = K_{h}(\Theta) x~,\quad \forall h \in [H]~,
\end{align*}
where the gain matrices $K_h(\Theta)$ are given by 
\begin{align}
\label{eq:K-matrix}
\!\!K_h(\Theta) \!=\! -(R \!+\! B^\top P_{h}(\Theta)B)^{-1} B^\top P_{h}(\Theta) A.
\end{align}
Here the symmetric positive semidefinite matrices $P_h(\Theta)$ are defined recursively by the \emph{Riccati iteration}
\begin{align}
    &P_{h}(\Theta) = Q + A^\top P_{h+1}(\Theta) A\label{eq:P-matrix}\\
     &\!-\!A^\top P_{h+1}(\Theta)B(R \!+\! B^\top P_{h+1}(\Theta)B)^{-1} B^\top P_{h+1}(\Theta) A \nonumber.
\end{align}
with $P_{H+1}(\Theta) := 0$.
The optimal cost is given by 
\begin{align}
\label{eq:opt_cost}
    \!\!\!\!\!J_h^*(\Theta,x) \!=\! x^\top\! P_{h}(\Theta)x \!+\!\!\! \sum\nolimits_{h'\!=h}^{H}\!\!\!\ex{w_{h'}^\top P_{h'+1}(\Theta) w_{h'}}\!.
\end{align}


\noindent
We let the agent play $K$ episodes and measure the performance by cumulative regret.\footnote{In the following, we add subscript $k$ to denote the variables for the $k$-th episode -- state $x_{k,h}$, control $u_{k,h}$, noise $w_{k,h}$ and cost $c_{k,h}$.} In particular, if the true system dynamics are $\Theta_* \!=\! [A_*, B_*]^\top$, the cumulative regret of the first $K$ episodes is given by
\begin{align}
\label{eq:regret}
    \cR(K) \!:=\! \sum\nolimits_{k=1}^K \left(J_1^{\pi_k}(\Theta_*,x_{k,1}) - J_1^*(\Theta_*,x_{k,1})\right),
\end{align}
where $J_1^*(\Theta_*,x_{k,1})$ is the (expected) cost under an optimal policy for episode $k$ (computed via~\eqref{eq:opt_cost}), and $J_1^{\pi_k}(\Theta_*,x_{k,1})$ is the (expected) cost under the chosen policy $\pi_k$ at the start of episode $k$ (computed via~\eqref{eq:pi_cost}). We seek to attain a sublinear regret $\cR(K) \!=\!o(K)$, which ensures that the agent finds the optimal policy as $K \!\to\! \infty$.
We end this section by presenting our assumptions on the LQ system \eqref{eq:model}, which are common in the LQ control literature \cite{wang2020episodic}.
\begin{assumption}[Boundedness] (a) The true system dynamics $\Theta_*$ is a member of a set $\cS\!:=\!\{\Theta \!=\! [A,B]^\top
\!:\! \norm{\Theta}_{\F} \!\leq\! 1 \text{ and } [A,B] \text{ is controllable}\}$. (b) There exist constants $C$, $C_A$, $C_B$ such that $\norm{A_*} \!\leq\! C_A \!<\! 1$, $\norm{B_*} \!\leq\! C_B \!<\!1$, and $\norm{Q} \!\leq\! C$, $\norm{R} \!\leq\! C$, (c) For all $k\!\geq\! 1$, $\norm{x_{k,1}} \!\leq\! 1$. (d) The noise  $w_{k,h}$ at any $k\!\geq\! 1$ and $h\!\in\![H]$, is (i) independent of all other randomness, (ii) $\ex{w_{k,h}} \!=\! 0$, 
    and (iii) $\norm{w_{k,h}}_2 \!\leq\! C_w \!<\!1$.
    (e) There exists a constant $\gamma$ such that $C_A + \gamma C_B +C_w \!\leq\! 1$.

\label{ass:regularity}
\end{assumption}

\subsection{Differential Privacy}

We now formally define the notion of differential privacy in the context of episodic LQ control. We write $v \!=\! (v_1, \ldots,v_K) \!\in\! \cV^K$ to denote a sequence of $K$ unique users participating in the private RL protocol with an RL agent $\cM$, where $\cV$ is the set of all users. Each user $v_k$ is identified by the state responses  $\lbrace x_{k,h+1} \rbrace_{h \in [H]}$ she gives to the controls $\lbrace u_{k,h}\rbrace_{h \in [H]}$ chosen by the agent. We write $\cM(v)\!=\!\lbrace u_{k,h} \rbrace_{k \in [K],h\in[H]} \in (\Real^d)^{KH}$ to denote the privatized controls chosen by the agent $\cM$ when interacting with the users $v$. 
Informally, we will be interested in randomized algorithms $\cM$ so that the knowledge of the output $\cM(v)$ and all but the $k$-th user $v_k$ does not reveal `much' about $v_k$. 
We formalize in the following definition, which is adapted from \cite{dwork2014algorithmic}.
\begin{definition}[Differential Privacy (DP)]
For any $\epsilon \!\geq\! 0$ and $\delta \!\in\! [0,1]$, an algorithm $\cM \!:\! \cV^K \!\ra\! (\Real^d)^{KH}$ is $(\epsilon,\delta)$-differentially private if for all $v,v' \in \cV^K$ differing on a single user and all subset of controls $\cU \!\!\subset\!\! \left(\Real^d\right)^{KH}$, 
\beqn
\prob{\cM(v)\in \cU} \leq \exp(\epsilon) \prob{\cM(v')\in \cU} + \delta~.
\eeqn
\label{def:DP}
\end{definition}
\vspace{-3mm}

We now relax this definition motivated by the fact that the controls recommended to a given user $v_k$ is only observed by her. We consider \emph{joint differential privacy} \cite{kearns2014mechanism}, which requires that simultaneously for all $k$, the
joint distribution on controls sent to users other than $v_k$ will not change substantially upon changing the state responses
of the user $v_k$. To this end, we let $\cM_{-k}(v)\!:=\!\cM(v)\!\setminus\! \lbrace u_{k,h} \rbrace_{h \in [H]}$ to denote all the controls chosen by the agent $\cM$ excluding those recommended to $v_k$.
\begin{definition}[Joint Differential Privacy (JDP)]
For any $\epsilon \!\geq\! 0$ and $\delta \!\in\! [0,1]$,
an algorithm $\cM \!:\! \cV^K \!\ra\! (\Real^d)^{KH}$ is $(\epsilon,\delta)$-jointly differentially private if for all $k \!\in\! [K]$, all $v,v' \!\in\! \cV$ differing on the $k$-th user and all subset of controls $\cU_{-k} \!\subset\! \left(\Real^d\right)^{(K\!-\!1)H}\!\!$ given to all but the $k$-th user, 
\beqn
\prob{\cM_{-k}(v)\in \cU_{-k}} \leq \exp(\epsilon) \prob{\cM_{-k}(v')\in \cU_{-k}} + \delta~.
\eeqn
\label{eq:JDP}
\end{definition}
\vspace{-3mm}

This relaxation is necessary
in our setting since knowledge of the controls recommended to the user $v_k$ can reveal a lot of information about her state responses. It weakens the constraint of DP only in that
the controls given specifically to $v_k$ may be sensitive in her state responses. However, it is
still a very strong definition since it protects $v_k$ from any arbitrary collusion of other users against her, so long as she does
not herself make the controls reported to her public.

In this work, we look for algorithms that are $(\epsilon,\delta)$-JDP. But, we will build our algorithm upon standard DP mechanisms. Furthermore, to establish privacy, we will use
a different relaxation
called \emph{concentrated differential privacy} (CDP) \cite{bun2016concentrated}. Roughly, a mechanism is CDP if the privacy loss has Gaussian tails.
To this end, we let $\cM$ to be a mechanism taking as input a data-stream $x \!\in\! \cX^n$ and releasing output from some range $\cY$.


\begin{definition}[Concentrated Differential Privacy (CDP)]
For any $\rho \geq 0$, an algorithm $\cM \!:\! \cX^n \!\to\! \cY$ is $\rho$-zero-concentrated differentially private if for all $x, x'\in \cX^n$ differing on a single entry and all $\alpha \in (1,\infty)$, 
\beqn
\D_{\alpha}\left(\cM(x)||\cM(x')\right) \leq \rho\;\alpha~,
\eeqn
where $\D_{\alpha}\left(\cM(x)||\cM(x')\right)$ is the $\alpha$-Renyi divergence between the distributions of $\cM(x)$ and $\cM(x')$.\footnote{For two probability distributions $P$ and $Q$ on $\Omega$, the $\alpha$-Renyi divergence $\D_{\alpha}(P||Q):=\frac{1}{\alpha-1}\ln \left(\int_{\Omega}P(x)^\alpha Q(x)^{1-\alpha}\diff \!x\right)$.}
\label{eq:zCDP}
\end{definition}

\vspace{-2mm}
\section{OFU-Based Control}

Our proposed private RL algorithm implements the optimism in the face of uncertainty (OFU) principle in LQ systems. As in~\cite{abbasi2011regret}, the key to implementing the OFU-based control is a high-probability confidence set for the unknown parameter matrix $\Theta_*$. 

\subsection{Adaptive Control with Changing
Regularizers}
\label{subsec:changing-regularizer}

We start with the adaptive LQ control with \emph{changing regularizers}. This not only allows us to generalize previous results for non-private control, but more importantly serves as a basis for the analysis of private control in the next section. We first define the following compact notations.
For a state and control pair at step $h$ in episode $k$, i.e., $x_{k,h}$ and $u_{k,h}$, we write $z_{k,h} \!=\! [x_{k,h}^\top, u_{k,h}^\top]^\top$. For any $k \!\geq\! 1$, we define the following matrices: $Z_k\!:=\![ z_{k',h'}^\top]_{k'\in[k-1],h'\in[H]}$, $X_k^{\text{next}}\!:=\![ x_{k',h'+1}^\top]_{k'\in[k-1],h'\in[H]}$ and $W_k\!:=\![ w_{k',h'}^\top]_{k'\in[k-1],h'\in[H]}$. 
For two matrices $A$ and $B$, we also define $\norm{A}_{B}^2 \!:=\! \text{trace}(A^{\top} B A)$.
Now, at every episode $k$, we consider the following ridge regression estimate w.r.t. a regularizing p.d. matrix $ H_k \in \Real^{(n+d)\times(n+d)}$:
\begin{align*}
    {\Theta}_k &:=\argmin_{\Theta \in \Real^{(n+d)\times n}} \norm{X_k^{\text{next}}-Z_k\Theta}_{\F}^2 + \norm{\Theta}_{H_k}^2\\
    &=(Z_k^{\top}Z_k + H_k)^{-1}Z_k^{\top}X_k^{\text{next}},
\end{align*}
In contrast to the standard online LQ control \cite{abbasi2011regret}, here the sequence of matrices $\lbrace Z_k^{\top}Z_k \rbrace_{k\geq 1}$ is perturbed by a sequence of regularizers $\lbrace H_k \rbrace_{k\geq 1}$. In particular, when $H_k \!=\! \lambda I$, we get back the standard estimate of~\cite{abbasi2011regret}. In addition, we also allow $Z_k^{\top}X_k^{\text{next}}$ to be perturbed by a matrix $L_k$ at every episode $k$. Setting $V_k\!:=\! Z_k^{\top}Z_k \!+\! H_k$ and $U_k\!:=\!Z_k^{\top}X_k^{\text{next}} \!+\! L_k$, we now define the estimate under changing regularizers $\{H_k\}_{k\ge 1}$ and $\{L_k\}_{k\ge 1}$ as
\begin{align}
\label{eq:lse-changing}
    {\hat \Theta}_k = {V}_k^{-1} {U}_k~.
\end{align}

\noindent
We make the following assumptions on the sequence of regularizers $\{H_k\}_{k\ge 1}$ and $\{L_k\}_{k\ge 1}$.
\begin{assumption}[Regularity]
\label{ass:regularizer}
For any $\alpha \in (0,1]$, there exist constants $\lambda_{\max}$, $\lambda_{\min}$ and $\nu$ depending on $\alpha$ such that, with probability at least $1-\alpha$, for all $k \in [K]$,
\begin{align*}
    \norm{H_k} \leq \lambda_{\max}, \;\; \norm{H_k^{-1}} \leq 1/\lambda_{\min} \;\; \text{and}\;\; \norm{L_k}_{H_k^{-1}} \leq \nu.
\end{align*}
\end{assumption}

\begin{lemma}[Concentration under changing regularizers]
\label{lem:CI}
Under assumptions \ref{ass:regularity} and \ref{ass:regularizer}, the following holds:
\beqn
\forall \alpha \!\in\! (0,1], \quad \prob{\exists k \!\in\! \Nat \!:\;\norm{\Theta_* \!-\! \hat\Theta_k}_{{V}_k} \!\geq\! \beta_{k}(\alpha)} \!\leq\! \alpha~,
\eeqn
where $ {\beta_{k}(\alpha)} \!:=\! C_w\!\sqrt{2\ln\!\big(\frac{2}{\alpha}\big)\! \!+\! n\ln\det\left(I\!+\!\lambda_{\min}^{-1}Z_k^\top Z_k\right)} \!+\! \sqrt{\lambda_{\max}} \!+\! \nu$.
\end{lemma}

\noindent
Lemma \ref{lem:CI} helps us to introduce the following high probability confidence set   
    \begin{align}
    \label{eq:Confidence_set}
        \mathcal{C}_k(\alpha):= \left\{ \Theta: \norm{\Theta - \hat\Theta_k}_{V_k} \leq \beta_{k}(\alpha)\right\}.
    \end{align}    
We then search for an optimistic estimate $\widetilde{\Theta}_k$ within this confidence region $\mathcal{C}_k(\alpha)$, such that
\begin{align}
\label{eq:OFU}
    \widetilde{\Theta}_k \in \argmin_{\Theta \in \mathcal{C}_k(\alpha) \cap \mathcal{S} } J_1^*(\Theta,x_{k,1}),
\end{align}
where $J_1^*(\Theta,x_{k,1})$ is the optimal cost when system dynamics are $\Theta$ (can be computed from~\eqref{eq:opt_cost}). With the estimate $\widetilde{\Theta}_k$, the agent then chooses policy $\pi_k$ and selects the controls recommended by this policy
\beq
\label{eq:controls}
u_{k,h} := \pi_{k,h}(x_{k,h})= K_h(\widetilde{\Theta}_k)x_{k,h}~,
\eeq
where $K_h(\widetilde{\Theta}_k)$ can be computed from~\eqref{eq:K-matrix}. We call this procedure \RL\ and bound its regret as follows. 


\begin{thm}[Regret under changing regularizers]
\label{thm:regret}
   Under Assumptions~\ref{ass:regularity} and~\ref{ass:regularizer}, for any $\alpha \!\in\! (0,1]$, with probability at least $1-\alpha$, the cumulative regret of \RL\ satisfies
\begin{align*}
    \mathcal{R}(K) 
    &= O\left(H\sqrt{K}\big(\sqrt{H}+ n(n+d)\psi_{\lambda_{\min}}+\ln(1/\alpha)\big) \right)\\
    +& O\left(H\sqrt{K} \left(\sqrt{\lambda_{\max}} + \nu\right) \sqrt{n(n+d)\psi_{\lambda_{\min}}} \right)~, 
\end{align*}
where $\psi_{\lambda_{\min}}\!:=\!\ln\left(1\!+\!HK/(n+d)\lambda_{\min}\right)$.
\end{thm}
\noindent
\textbf{Proof sketch.} Inspired by~\cite{wang2020episodic}, we first decompose the regret under the following `good' event: $\mathcal{E}_K(\alpha) \!:=\! \{\Theta_* \!\in\! \mathcal{C}_k(\alpha) \!\cap\! \mathcal{S}, \forall k \!\in\![K] \},$
which, by Assumption \ref{ass:regularity} and Lemma \ref{lem:CI}, holds w.p. at least $1\!-\!\alpha$. Then, under the `good' event, the cumulative regret (\ref{eq:regret}) can be written as 
    \begin{align*}
        \mathcal{R}(K) \leq \sum\nolimits_{k=1}^K\sum\nolimits_{h=1}^{H}(\Delta_{k,h} + \Delta_{k,h}^{\prime} + \Delta_{k,h}^{\prime\prime}), \;\; \text{where}
    \end{align*}
    \begin{align*}
        &\Delta_{k,h} \!:=\!\mathbb{E}\left[J_{h+1}^{\pi_k}(\Theta_*,x_{k,h+1}) \!\!\mid\! \mathcal{F}_{k,h}\right] \!-\! J_{h+1}^{\pi_k}(\Theta_*,x_{k,h+1}),\\
        &\Delta_{k,h}^{\prime}\!:=\! \norm{x_{k,h+1}}_{\widetilde{P}_{k,h+1}} \!\!\!-\! \mathbb{E}\left[\norm{x_{k,h+1}}_{\widetilde{P}_{k,h+1}} \!\!\mid\! \mathcal{F}_{k,h}\right]\; \text{and}\\
        &\Delta_{k,h}^{\prime\prime} \!:=\! \norm{\Theta_*^\top z_{k,h}}_{\tp_{k,h+1}} \!-\! \norm{\widetilde{\Theta}_k^\top z_{k,h}}_{\tp_{k,h+1}},
    \end{align*}
    in which $\widetilde{P}_{k,h} \!:=\! P_h(\widetilde{\Theta}_k)$ is given by \eqref{eq:P-matrix} and $\mathcal{F}_{k,h}$ denotes all the randomness present before time $(k,h)$. 
    
    Now, we are going to bound each term, respectively. For the first two terms, we can show that both of them are bounded martingale difference sequences. Therefore, by Azuma–Hoeffding inequality, we have $ \sum_{k,h} \Delta_{k,h} \!=\! O\big(\sqrt{KH^3}\big)$ and $ \sum_{k,h} \Delta'_{k,h} \!=\! O\big(\sqrt{KH}\big)$ with high probability.
     We use Lemma \ref{lem:CI} and the OFU principle \eqref{eq:OFU} to bound the third term as
     $\sum_{k,h}\Delta''_{k,h} \!=\! O\big(H\sqrt{K}\beta_k(\alpha) \sqrt{\ln\det\left(I\!+\!\lambda_{\min}^{-1}Z_k^\top Z_k\right)} \big)$.
     To put everything together, first note from Assumption \ref{ass:regularity} that 
    \begin{align*}
    \ln\det\left(I\!+\!\lambda_{\min}^{-1}Z_k^\top Z_k\right) \!\leq\! (n\!+\!d)\ln\left(1\! +\! \frac{HK(1+\gamma)^2}{(n+d)\lambda_{\min}}\right).
    \end{align*}
    Plugging this into $\beta_k(\alpha)$ given in Lemma~\ref{lem:CI} and the third term above, yields the final result. \qed
    


We end the section with a proof sketch of Lemma \ref{lem:CI}.

\noindent
\textbf{Proof sketch (Lemma \ref{lem:CI}).} Under Assumptions \ref{ass:regularity} and \ref{ass:regularizer}, with some basic algebra, we first have 
\begin{align*}
    &\norm{{\Theta}_* - \hat\Theta_k}_{{V}_k}
     = \norm{H_k\Theta_* - Z_k^\top W_k -L_k}_{{V}_k^{-1}}\\
     \leq & \underbrace{\norm{Z_k^\top W_k}_{(Z_k^\top Z_k + \lambda_{\min}I)^{-1}}}_{\mathcal{T}_1} + \underbrace{\norm{H_k^{\frac{1}{2}}}_2+ \norm{L_k}_{H_k^{-1}}}_{\mathcal{T}_2}.
\end{align*}
By Assumption~\ref{ass:regularizer}, we have w.p. at least $1-\alpha$, $\mathcal{T}_2 \leq \sqrt{\lambda_{\max}} + \nu$. To bound $\mathcal{T}_1$, by the boundedness of $w_{k,h}$ in Assumption~\ref{ass:regularity}, we first note that each row of the matrix $W_k$ is a sub-Gaussian random vector with parameter $C_w$. We then generalize the self-normalized concentration inequality of vector-valued martingales \cite[Theorem 1]{abbasi2011improved} to the setting of matrix-valued martingales. In particular, we show that w.p. at least $1-\alpha$, 
\begin{align*}
          \mathcal{T}_1 \leq C_w \sqrt{2\ln \left(1/\alpha\right)+n\ln\det(I\!+\!\lambda_{\min}^{-1}Z_k^\top Z_k) }.
\end{align*}
Combining the bounds on $\mathcal{T}_1$ and $\mathcal{T}_2$ using a union bound argument, yields the final result. \qed

\subsection{Private Control}

In this section, we introduce the \PRL\ algorithm (Alg. \ref{alg:RL}). At every episode $k$, we keep track of the history via private version of the matrices $Z_k^\top Z_k$ and $Z_k^\top X_k^{\text{next}}$. To do so, we first initialize two private counter mechanisms $\cB_1$ and $\cB_2$, which take as parameters the privacy levels $\epsilon$, $\delta$, number of episodes $K$, horizon $H$ and a problem-specific constant $\gamma$ (see Assumption \ref{ass:regularity}). The counter $\cB_1$ (resp. $\cB_2$) take as input an event stream of matrices $\lbrace \sum_{h=1}^{H} z_{k,h}z_{k,h}^\top\rbrace_{k \in [K]}$ (resp. $\lbrace \sum_{h=1}^{H} z_{k,h}x_{k,h+1}^\top\rbrace_{k \in [K]}$), and at the start of each episode $k$, release the private version of the matrix $Z_k^\top Z_k$ (resp. $Z_k^\top X_k^{\text{next}}$), which itself is a matrix of the same dimension. Let $T_{1,k}$ and $T_{2,k}$ denote the privatized versions for $Z_k^\top Z_k$ and $Z_k^\top X_k^{\text{next}}$, respectively. For some $\eta > 0$ (will be determined later), we define $V_k\!:=\!T_{1,k} \!+\! \eta I$ and $U_k\!:=\!T_{2,k}$. We now instantiate the general \RL\ procedure under changing regularizers (Section \ref{subsec:changing-regularizer}) with these private statistics. First, we compute the point estimate $\hat \Theta_k$ from (\ref{eq:lse-changing}) and build the confidence set $\cC_k(\alpha)$ from (\ref{eq:Confidence_set}). Then, we choose the most optimisitic policy $\pi_k$ w.r.t. the entire set $\cC_k(\alpha)$ from (\ref{eq:OFU}) and (\ref{eq:controls}). Finally, we execute the policy for the entire episode and update the counters with observed trajectory.

We now describe the private counters $\cB_1$ adapting the \emph{Binary counting mechanism} of \cite{chan2011private}. First, we write $\Sigma_1[i,j]=\sum_{k=i}^{j}\sum_{h=1}^{H}z_{k,h}z_{k,h}^\top$ to denote a partial sum (\psum) involving the state-control pairs in episodes $i$ through $j$. Next, we consider a binary interval tree, where each leaf node represents an episode (i.e., the tree has $k-1$ leaf nodes at the start of episode $k$), and each interior node represents the range of episodes covered by its children. At the start of episode $k$, we first release a noisy \psum\ $\hat \Sigma_1[i,j]$ corresponding to each node in the tree. Here $\hat \Sigma_1[i,j]$ is obtained by perturbing both $(p,q)$-th and $(q,p)$-th, $1 \!\leq\!p \!\leq\! q \!\leq\! (n\!+\!d)$, entries of $\Sigma_1[i,j]$ with i.i.d. Gaussian noise $\zeta_{p,q}\!\sim\!\cN(0,\sigma_1^2)$.\footnote{This will ensure symmetry of the \psums\ even after adding noise.} Then $T_{1,k}$ is computed by summing up the noisy \psums\ released by the set of nodes that uniquely cover the range $[1,k\!-\!1]$. Observe that, at the end each episode, the mechanism only needs to store noisy \psums\ required for computing private statistics at future episodes, and can safely discard \psums\ that are
no longer needed. For the private counter $\cB_2$, we maintain \psums\ $\Sigma_2[i,j]\!=\!\sum_{k=i}^{j}\sum_{h=1}^{H}z_{k,h}x_{k,h+1}^\top$ with i.i.d. noise $\cN(0,\sigma_2^2)$ and compute the private statistics $T_{2,k}$ using a similar procedure. The noise levels $\sigma_1$ and $\sigma_2$ depends on the problem intrinsics ($K$, $H$, $\gamma$) and privacy parameters ($\epsilon$, $\delta$). These, in turn, govern the constants $\lambda_{\max}$, $\lambda_{\min}$, $\nu$ appearing in the confidence set $\cC_k(\alpha)$ and the regularizer $\eta$. The details will be specified in the next Section as needed.



\setlength{\textfloatsep}{0pt}
\begin{algorithm}[t!]
\caption{\PRL}
\label{alg:RL}
\DontPrintSemicolon
\KwIn{Number of episodes $K$, horizon $H$, privacy level $\epsilon \!>\! 0$, $\delta \!\in\! (0,1]$, constants $\gamma$, $C_w$, confidence level $\alpha \in (0,1]$}
initialize private counters $\cB_1$ and $\cB_2$ with parameters $K,H,\epsilon,\delta,\gamma$\;
\For{each episode $k\!=\!1,2,3,\ldots,K$}{
compute private statistics $T_{1,k}$ and $T_{2,k}$\;
construct confidence set $\cC_k(\alpha)$ \;
find $\widetilde{\Theta}_k \in \argmin_{\Theta \in \mathcal{C}_k(\alpha) \cap \mathcal{S} } J_1^*(\Theta,x_{k,1})$\;
\For{each step $h\!=\!1,2,\ldots,H$}{
execute control $u_{k,h}\!=\!K_h(\widetilde{\Theta}_k)x_{k,h}$\;
observe cost $c_{k,h}$ and next state $x_{k,h+1}$\;
}
send $\sum_{h=1}^{H}z_{k,h}z_{k,h}^\top$ and $\sum_{h=1}^{H}z_{k,h}x_{k,h+1}^\top$ to the counters $\cB_1$ and $\cB_2$, respectively
}
\end{algorithm}


\section{Privacy and Regret Guarantees}

In this section, we show that \PRL\ is a JDP algorithm with sublinear regret guarantee.

\subsection{Privacy Guarantee}\label{subsec:privacy}

\begin{thm}[Privacy]
Under Assumption \ref{ass:regularity}, for any $\epsilon \!>\! 0$ and $\delta \!\in\! (0,1]$, \PRL\ is $(\epsilon,\delta)$-JDP.
\label{thm:privacy}
\end{thm}
\noindent
\textbf{Proof sketch.}
We first show that both the counters $\cB_1$ and $\cB_2$ are $(\epsilon/2,\delta/2)$-DP. We begin with the counter $\cB_1$. To this end, we need to determine a global upper bound $\Delta_1$ over the $\LL_2$-sensitivity of all the \psums\ $\Sigma_1[i,j]$. Informally, $\Delta_1$ encodes the maximum change in the Frobenious norm of each \psum\ if the trajectory of a single episode is changed. By Assumption \ref{ass:regularity}, we have $\norm{z_{k,h}} \!\leq\! 1\!+\!\gamma$, and hence $\Delta_1\!=\!H(1\!+\!\gamma)^2$. Since the noisy \psums\ $\hat\Sigma_1[i,j]$ are obtained via Gaussian mechanism, we have that each $\hat\Sigma_1[i,j]$ is $(\Delta_1^2/2\sigma_1^2)$-CDP \cite[Proposition 1.6]{bun2016concentrated}.
We now see that every episode appears only in at most $m\!:=\! \lceil \log_2 K \rceil$ \psums\ $\Sigma_1[i,j]$. Therefore, by the composition property, the whole counter $\cB_1$ is $(m\Delta_1^2/2\sigma_1^2)$-CDP, and thus, in turn, $\left(\frac{m\Delta_1^2}{2\sigma_1^2}\!+\!2\sqrt{\frac{m\Delta_1^2}{2\sigma_1^2}\ln\left(\frac{2}{\delta}\right)},\frac{\delta}{2}\right)$-DP for any $\delta \!>\! 0$ \cite[Lemma 3.5]{bun2016concentrated}. Now, setting $\sigma_1^2 \!\approx\!8m\Delta_1^2\ln(2/\delta) / \epsilon^2$, we can ensure that $\cB_1$ is $(\epsilon/2,\delta/2)$-DP. A similar analysis yields that counter $\cB_2$ is $(\epsilon/2,\delta/2)$-DP if we set  $\sigma_2^2\!\approx\!8m\Delta_2^2\ln(2/\delta) / \epsilon^2$, where $\Delta_2 \!:=\! H(1\!+\!\gamma)$.

To prove Theorem \ref{thm:privacy}, we now use the \emph{billboard lemma} \cite[Lemma 9]{hsu2016private} which, informally, states that an algorithm is JDP under continual observation if the output sent to each user is a function of the user’s private data and a common quantity computed using standard differential privacy. Note that at each episode $k$, \PRL\ computes private statistics $T_{1,k}$ and $T_{2,k}$ for all users using the counters $\cB_1$ and $\cB_2$. These statistics are then used to compute the policy $\pi_k$. By composition and post-processing properties of DP, we can argue that the sequence of policies $\lbrace \pi_k \rbrace_{k \in [K]}$ are computed using an $(\epsilon,\delta)$-DP mechanism.
Now, the controls $\lbrace u_{k,h} \rbrace_{h \in [H]}$ during episode $k$ are generated using the policy $\pi_k$ and the user's private data $x_{k,h}$ as $u_{k,h}\!=\!\pi_{k,h}(x_{k,h})$. Then, by the billboard lemma, the composition of the controls $\lbrace u_{k,h} \rbrace_{k \in [K],h\in [H]}$ sent to all the users is $(\epsilon,\delta)$-JDP.\qed


\subsection{Regret Guarantee}

\begin{thm}[Private regret] Under Assumption \ref{ass:regularity}, for any privacy parameters $\epsilon \!>\! 0$ and $\delta \!\in\! (0,1]$, and for any $\alpha \!\in\!(0,1]$, with probability at least $1\!-\!\alpha$, \PRL\ enjoys the regret bound
\begin{align*}
  &\cR(K)= O \!\left(H^{3/2}\sqrt{K} \big( n(n\!+\!d) \ln K \!+\! \ln(1/\alpha) \big)\right)\\
&\!+\!O \!\left(\!H^{3/2}\!\sqrt{K}  \ln K\!\left(n(n\!+\!d)\!+\!\!\sqrt{\ln K/\alpha}\right)\frac{\ln(1/\delta)^{1/4}}{\epsilon^{1/2}} \right)\!.
\end{align*}
\label{thm:private-regret}
\end{thm}
\vspace{-2mm}
\noindent
Theorems \ref{thm:privacy} and \ref{thm:private-regret} together imply that \PRL\ can achieve a sub-linear regret under $(\epsilon,\delta)$-JDP privacy guarantee. Furthermore, comparing Theorem \ref{thm:private-regret} with Theorem \ref{thm:regret}, we see that the first term in the regret bound corresponds to the non-private regret, and the second term is the cost of privacy. More importantly, the cost due to privacy grows only as $\frac{\ln(1/\delta)^{1/4}}{\epsilon^{1/2}}$ with $\epsilon,\delta$.

\vspace{0.25mm}
\noindent
\textbf{Proof sketch (Theorem \ref{thm:private-regret}).}
First note that the private statistics $T_{1,k}$ can be computed by summing at most $m\!=\!\lceil \log_2 K \rceil$ noisy \psums\ $\hat \Sigma_1[i,j]$. We then have that the total noise $N_k$ in each $T_{1,k}$ is a symmetric matrix with it's $(p,q)$-th entry, $1 \!\leq\! p \!\leq\! q \!\leq\! (n\!+\!d)$, being i.i.d. $\cN(0,m\sigma_1^2)$. Therefore, by an adaptation of~\cite[Corollary 4.4.8]{vershynin2018high}, we have w.p. at least $1\!-\!\alpha/2K$,
\begin{align*}
    \norm{N_{k} } \!\leq\! \Lambda\!:=\! \sigma_1\sqrt{m} \left(4\sqrt{n\!+\!d}\!+\!\sqrt{8\ln(4K/\alpha)}\right).
\end{align*}
Similarly, the total noise $L_k$ in each $T_{2,k}$ is an $(n+d)\times n$ matrix, whose each entry is i.i.d. $\cN(0,m\sigma_2^2)$. Hence $\norm{L_k}_{\F}^2/m\sigma_2^2$ is a $\chi^2$-statistic with $n(n\!+\!d)$ degrees of freedom, and therefore, by~\cite[Lemma 1]{laurent2000adaptive}, we have w.p. at least $1\!-\!\alpha/2K$,
\begin{align*}
    \!\norm{L_k}_{\F} \!\leq\! \sigma_2\sqrt{m}\left(\!\sqrt{2n(n\!+\!d)} \!+\! \sqrt{4\ln(2K/\alpha)} \right).
\end{align*}
By construction, we have the regularizer $H_k\!=\!N_k\!+\!\eta I$. Setting $\eta\!=\!2\Lambda$, we ensure that $H_k$ is p.d., and hence $\norm{L_k}_{H_k^{-1}} \!\leq\! \Lambda^{-1/2} \norm{L_k}_{\F}$. Then, by a union bound argument, Assumption \ref{ass:regularizer} holds for $\lambda_{\min}\!=\!\Lambda$, $\lambda_{\max}\!=\!3\Lambda$ and $\nu \!=\! \sigma_2\sqrt{m/\Lambda}\left(\sqrt{2n(n\!+\!d)}\!+\!\!\sqrt{4\ln (2K/\alpha)} \right)$. Appropriating noise levels $\sigma_1,\sigma_2$ from Section \ref{subsec:privacy}, the regret bound now follows from Theorem \ref{thm:regret}.\qed

\vspace{-2.5mm}
\section{Conclusion}
\vspace{-1mm}
We develop the first DP algorithm, \PRL, for episodic LQ control. Through the notion of JDP, we show that it can protect private user information from being inferred by observing the control policy without losing much on its regret performance. We leave as future work private control of non-linear systems \cite{chowdhury2019online}.



\newpage
\enlargethispage{-1.2cm} 

\bibliographystyle{IEEEtran}
\bibliography{references}


\end{document}